\documentclass[runningheads]{llncs}

\usepackage{eccv}

\usepackage{eccvabbrv}
\usepackage{multirow}
\usepackage{graphicx}
\usepackage{makecell}

\usepackage{graphicx}
\usepackage{booktabs}
\usepackage{amsmath}
\usepackage[bb=dsserif]{mathalpha}
\usepackage{pifont}
\usepackage{amssymb}
\usepackage{xcolor}

\definecolor{mygreen}{rgb}{0.73,0.86,0.63}
\definecolor{mypurple}{rgb}{0.67,0.44,0.75}

\usepackage[accsupp]{axessibility}  

\usepackage[colorlinks,citecolor=eccvblue]{hyperref}
\usepackage{fontawesome5}
\usepackage[dvipsnames]{xcolor}
\definecolor{customgreen}{HTML}{95CA6D}
\definecolor{custompurple}{HTML}{B266FF}

\begin{document}

\title{AFF-ttention! Affordances and Attention models for Short-Term Object Interaction Anticipation} 

\titlerunning{AFF-ttention! Affordances and Attention models for STA}

\author{Lorenzo Mur-Labadia\inst{1} \and 
Ruben Martinez-Cantin\inst{1} \and
Jose J.Guerrero\inst{1} \and
Giovanni Maria Farinella \inst{2} \and
Antonino Furnari \inst{2}}

\authorrunning{L.Mur-Labadia et al.}

\institute{University of Zaragoza, Spain \\ \email{\{lmur, rmcantin, jguerrer\}@unizar.es} \and
University of Catania, Italy \\
\email{\{giovanni.farinella, antonino.furnari\}@unict.it}}
\maketitle

\begin{abstract}

Short-Term object-interaction Anticipation (STA) consists of detecting the location of the next-active objects, the noun and verb categories of the interaction, and the time to contact from the observation of egocentric video.
This ability is fundamental for wearable assistants or human-robot interaction to understand the user's goals, but there is still room for improvement to perform STA in a precise and reliable way.
In this work, we improve the performance of STA predictions with two contributions: 1) We propose STAformer, a novel attention-based architecture integrating frame-guided temporal pooling, dual image-video attention, and multiscale feature fusion to support STA predictions from an image-input video pair; 2) We introduce two novel modules to ground STA predictions on human behavior by modeling affordances.

First, we integrate an environment affordance model which acts as a persistent memory of interactions that can take place in a given physical scene.
Second, we predict interaction hotspots from the observation of hands and object trajectories, increasing confidence in STA predictions localized around the hotspot.
Our results show significant relative Overall Top-5 mAP improvements of up to $+45\%$ on Ego4D and $+42\%$ on a novel set of curated EPIC-Kitchens STA labels.
 \href{https://github.com/lmur98/AFFttention}{We will release the code, annotations, and pre-extracted affordances} on Ego4D and EPIC-Kitchens to encourage future research in this area. 
\end{abstract}

\section{Introduction}
\label{sec:intro}

\begin{figure}[t]
\centering
\includegraphics[width=\textwidth]{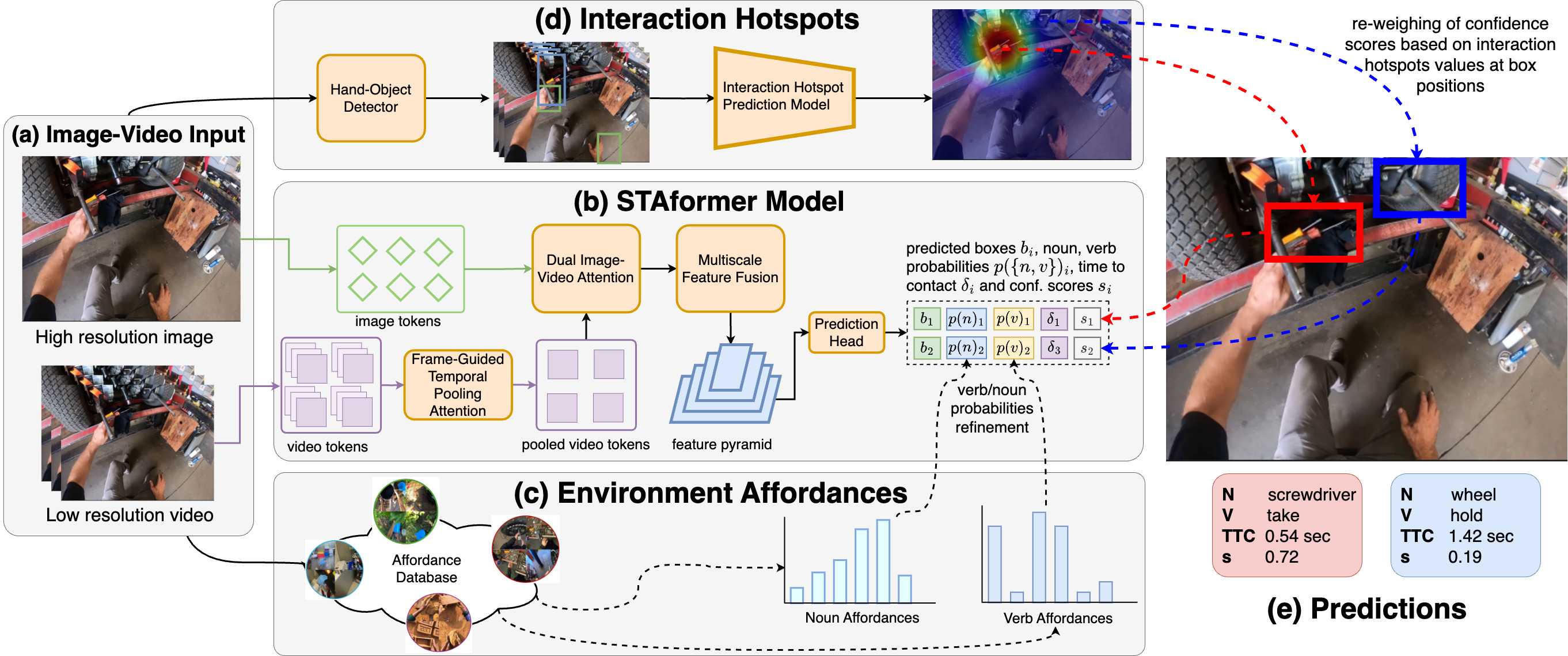}
\caption{(a) Our approach takes as input an image-video pair. (b) The input is processed by the proposed STAformer model which predicts object bounding boxes, the associated verb/noun probabilities, time-to-contact estimates and confidence scores. (c) Environment affordances are inferred from video and used to refine the predicted noun/verb probabilities. (d) Our model observes detected hand-object interactions in the video and predicts an interaction hotspot probability map, which is used to re-weigh confidence scores based on box locations, leading to (e) our final predictions.}
\label{fig:teaser}
\end{figure}


Anticipating the future is a fundamental ability for assistive egocentric devices and to support human-robot interaction. For example, a smart wearable device could alert an electrical operator before they short-circuit a switchboard, or a home robot can support the human by turning on appliances or moving objects according to their forecasted long-term goal. 
Predicting the future state of the scene from egocentric visual observations is a growing research area~\cite{plizzari2023outlook, rodin2021predicting}, 
with works tackling action anticipation~\cite{roy2024interaction, furnari2020rolling, chi2023adamsformer, nawhal2022rethinking, girdhar2021anticipative, zhong2023anticipative, zatsarynna2021multi}, locomotion prediction~\cite{lee2012discovering, park2016egocentric, bi2020can, marchetti2020multiple, kitani2012activity}, hands trajectory forecasting~\cite{liu2020forecasting, liu2022joint, bao2023uncertainty}, and next-active object detection~\cite{furnari2017next,ragusa2021meccano,jiang2021predicting,dessalene2021forecasting}.
Recently, Grauman et al.~\cite{grauman2022ego4d} defined the Short-Term Object Interaction Anticipation (STA) task as 
the simultaneous prediction of the action and object category, the object's bounding box, and the time to contact, and introduced an international challenge within the forecasting benchmark of the Ego4D dataset. Inspired by this challenge, the community proposed different approaches~\cite{chen2022internvideo, tong2022videomae, pasca2023summarize, ragusa2023stillfast, thakur2023enhancing, thakur2023guided, thakur2024leveraging}. Despite the progress in the area, our results show a large advantage over previous results, which highlights the room for improvement in accuracy and robustness.

%
Our aim with this work is to advance research in STA with two main contributions. First, 
we propose STAformer, a principled architecture unifying the computation of image and video inputs with attention-based components (Figure~\ref{fig:teaser}(a)-(b)). Differently from previous approaches~\cite{grauman2022ego4d,thakur2023guided,pasca2023summarize}, we explicitly designed STAformer to operate on an image-video input pair, which is specific to the considered STA task. Our architecture is a significant departure from convolutional baselines~\cite{grauman2022ego4d,ragusa2023stillfast} and aims to offer the convenience and state-of-the-art performance of attention-based feature extractors~\cite{oquab2023dinov2,bertasius2021space} and components~\cite{vaswani2017attention}.
Second, to tackle the challenges associated with relating past visual observations to future events from video, we propose two effective modules designed to ground predictions into human behavior by modeling affordances.
As highlighted in recent studies~\cite{plizzari2023can},  human activities exhibit consistency in similar environments. Hence, we first leverage environment affordances~\cite{nagarajan2019grounded}, estimated by matching the input observation to a learned affordance database, to predict probability distributions over nouns and verbs, which are used to refine verb and noun probabilities predicted by STAformer (Figure~\ref{fig:teaser}(c)). Our intuition is that linking a zone across similar environments captures a description of the feasible interactions, grounding predictions into previously observed human behavior.
The second affordance module aims to relate STA predictions to a spatial prior of where an interaction may take place in the current frame. This is done by predicting an interaction hotspot~\cite{liu2022joint}, which is used to re-weigh confidence scores of STA predictions depending on the object's locations (Figure~\ref{fig:teaser}(e)).

Experiments on Ego4D~\cite{grauman2022ego4d} and a novel set of curated STA annotations on the EPIC-Kitchens dataset~\cite{damen2018scaling} highlight the effectiveness of the proposed approach, which obtains significant relative improvements of $+45\%$ on the validation set of Ego4D v1, $+42.1\%$ on the validation set of Ego4D v2, $+30.3\%$ on the private test set of Ego4D, and $+42\%$ on EPIC-Kitchens, when measured with the official overall Top-5 mAP evaluation measure. The proposed approach currently scores first on the Ego4D Short-Term object-interaction Anticipation leaderboard.\footnote{See the supplementary material for more details.} Experiments also highlight the individual contributions of STAformer and the proposed modules to exploit affordances for STA prediction. In sum, the contributions of our work are as follows: 1) We introduce STAformer, a novel attention-based architecture specifically designed to process an input image-video pair, which achieves state-of-the-art performance on the two challenging Ego4D and EPIC-Kitchens benchmarks. 2) We propose two modules to ground STA predictions to human behavior by modeling environment affordances and interaction hotspots. The two modules are shown to be effective when coupled with STAformer, as well as previous architectures, which highlights the general usefulness of the approach. 3) We contribute a novel set of STA annotations, curated from public EPIC-Kitchens labels. This effectively provides the research community with a second large-scale and challenging benchmark for the STA task, besides the popular Ego4D.
We will publicly release the open-source implementation of STAformer and the affordance modules, the proposed affordance databases pre-computed on Ego4D and EPIC-Kitchens, and the novel EPIC-Kitchens STA annotations.
\section{Related works}


\textbf{Short-term Object Interaction Anticipation:}
Among seminal works, Furnari et al.~\cite{furnari2017next} initially introduced the concept of Next-Active Objects (NAO), proposing to detect future interacted objects by analyzing their trajectories as observed from the first-person point of view. 
Differently from action anticipation~\cite{damen2018scaling}, the NAO detection task is designed to provide grounded predictions in the form of bounding boxes, which can be particularly informative for wearable AI assistants or embodied robotic agents.
Unlike traditional object detection~\cite{girshick2015fast}, NAO prediction requires the ability to model the dynamics of the scene and anticipate the user's intention. 
Jiang et al.~\cite{jiang2021predicting} developed a method to predict the next-active object location in the form of a Gaussian heatmap from a single RGB image, combining visual attention with probabilistic maps of hand locations. Ego-OMG~\cite{dessalene2021forecasting} segments the NAO and predicts the interaction time using a contact anticipation map that captures scene dynamics. 
While previous works considered different task formulations and evaluation approaches, Grauman et al.~\cite{grauman2022ego4d} formalized NAO prediction by introducing the STA task and an associated challenge on the EGO4D dataset~\cite{grauman2022ego4d}. The initial baseline is composed of a Faster R-CNN branch to detect objects~\cite{girshick2015fast} and a SlowFast 3D CNN~\cite{feichtenhofer2019slowfast} for video processing.
Subsequent research introduced architectural enhancements and alternative approaches. Chen et al.~\cite{chen2022internvideo} employed pre-computed object detections using a DETR model and substituted SlowFast with a VideoMAE pre-trained ViT~\cite{tong2022videomae}. Pasca et al.~\cite{pasca2023summarize} proposed TransFusion, which employs a language encoder for action context summary, performing multi-modal fusion with visual features.
While previous works leveraged pre-extracted object detections for 2D image understanding, Ragusa et al.~\cite{ragusa2023stillfast} introduced StillFast, an end-to-end framework unifying the processing of 2D images and video in a combined backbone.
Thakur et al.~\cite{thakur2023enhancing} proposed GANO, an end-to-end model based on a transformer architecture including a novel guided attention mechanism. 
Guided attention was integrated within a StillFast architecture in~\cite{thakur2023guided}, achieving state-of-the-art results. Thakur et al.~\cite{thakur2024leveraging} introduced NAOGAT, a multi-modal transformer that attends detected objects and includes a motion decoder to track object trajectories. Despite the progress, previous works show incremental results over the baselines, indicating significant potential improvement.



\noindent
\textbf{Affordances for Anticipation:}
Defined by Gibson~\cite{gibson1977theory}, affordances are \textit{the potential actions that the environment offers to the agent}. 
The computational perception of affordances has been investigated in different forms. 
A line of works predicts affordance labels of object parts, requiring strong supervision in the form of manually annotated masks~\cite{mur2023bayesian, do2018affordancenet, myers2015affordance, nguyen2017object}. However, these methods are not ``grounded'' in human behavior as the annotator declares interaction regions outside of any interaction context~\cite{nagarajan2019grounded}.
Other works considered the problem of grounding affordance regions in images by leveraging videos depicting human-object interactions in a weakly supervised way, where only the action label is used as supervision without spatial annotations~\cite{nagarajan2019grounded,luo2023learning,goyal2022human,li2023locate}. Nagarajan et al.~\cite{nagarajan2019grounded} introduced the concept of ``interaction hotspots'' as the potential spatial regions where the action can occur.
Mur-Labadia et al.~\cite{mur2023multi} create a 3D multi-label mapping of affordances extracted from egocentric video.
Another line of work infers interaction hotspots from video by forecasting future hand movements to select candidate regions for future interactions~\cite{jiang2021predicting, liu2022joint, liu2020forecasting, goyal2022human}. 
Few works studied scene affordances to predict a list of likely actions that can be performed in a given scene~\cite{rhinehart2016learning,nagarajan2020ego}. In particular, Nagarajan et al.~\cite{nagarajan2020ego} proposed EGO-TOPO, a procedure to decompose a set of egocentric videos into a topological map encoding scene affordances.
Despite the interest in affordances, only a few works investigated how to exploit them for future predictions. Montesano et al.~\cite{montesano2008learning} predicted affordance effects for human-robot interaction. Koppula et al.~\cite{koppula2015anticipating} used object affordances to anticipate human behavior in the form of motion trajectories of objects and humans.
Nagarajan et al.~\cite{nagarajan2020ego} showed how scene affordances learned from egocentric video can improve long-term action anticipation.
Liu et al.~\cite{liu2020forecasting} tackled action anticipation by jointly predicting egocentric hand motion, interaction hotspots, and future actions.
Liu et al.~\cite{liu2022joint} highlighted how interaction hotspots predicted by forecasting hand motion can support action anticipation.
\section{STAformer Architecture for Short-Term Anticipation}
\begin{figure}[t]
\centering
\includegraphics[width=\textwidth]{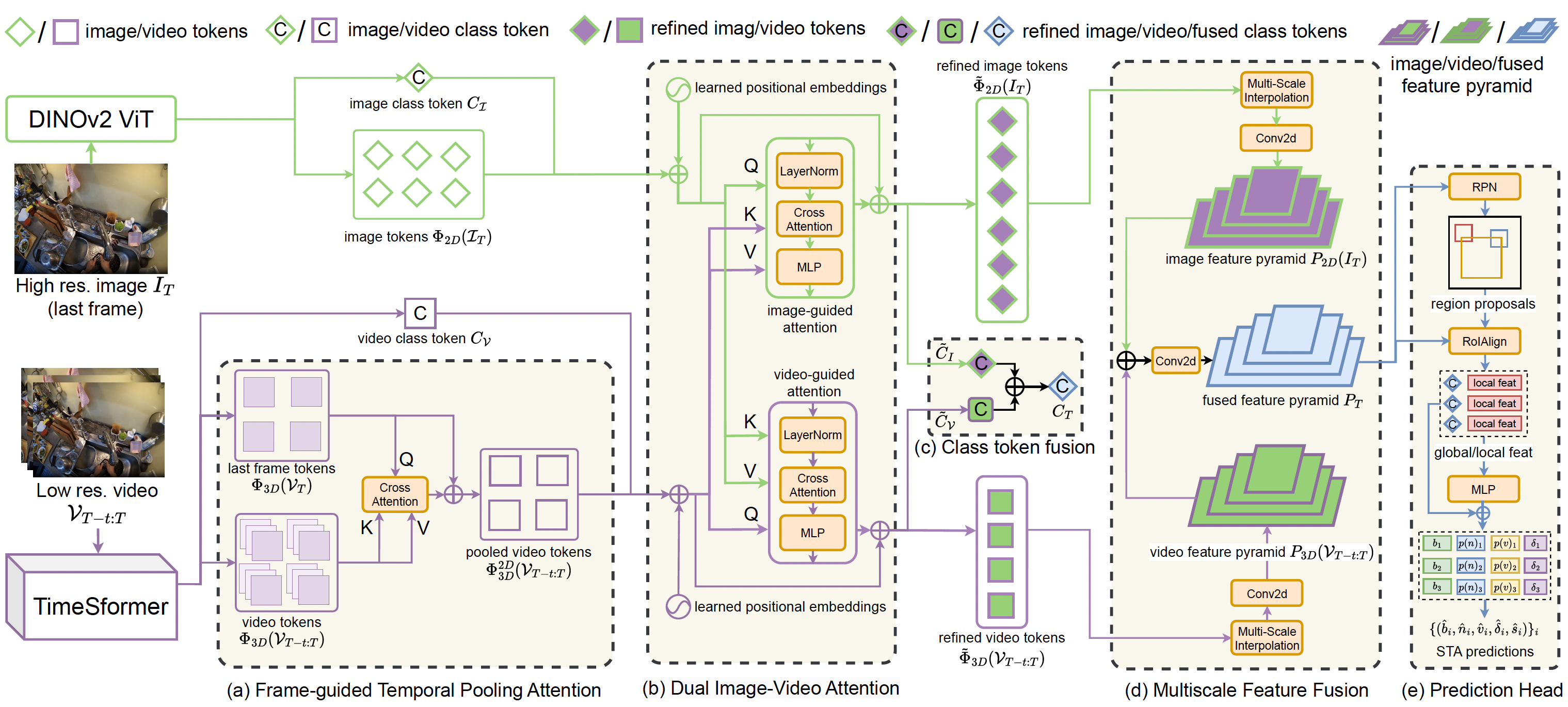}
\caption{\textbf{STAformer architecture.} DINO-v2 and TimeSformer extract 2D and 3D features form the image-video input. (a) Frame-guided temporal pooling attention spatially aligns video to image features. (b) Dual image-video attention enriches 2D features with temporal dynamics and 3D features with fine-grained image details. Image and video representations are joined to obtain a global class token (c) and a feature pyramid (d), from which we obtain the STA predictions (e).} 
\label{fig:encoder}
\end{figure}

As defined in \cite{grauman2022ego4d}, the goal of Short-Term object interaction Anticipation (STA) is to detect the Next-Active Object (NAO) from the observation of a given input video $\mathcal{V}_{:T}$ up to timestamp $T$. The model prediction are a set of detections, defined by the tuple $(b_i,n_i,v_i,\delta_i, s_i)$, denoting future interacted objects in the last observed frame $I_T$. Each bounding box $b_i$ is associated with an object category label $n_i$ (noun), a verb label indicating the interaction mode $v_i$, a time-to-contact $\delta_i$ indicating that the interaction will take place at time $T+\delta_i$, and a confidence score $s_i$.
We propose STAformer, a novel architecture that leverages pre-trained transformer models for image and video feature extraction~\cite{oquab2023dinov2,pramanick2023egovlpv2} and introduces novel attention-based components for image-video representation fusion. We illustrate the architecture in Figure~\ref{fig:encoder} and discuss it in the following. 

\vspace{1mm}
\noindent
\textbf{Feature Extraction:}
We follow previous work~\cite{grauman2022ego4d,ragusa2023stillfast} and process a high resolution image $I_T \in \mathbb{R}^{h_s \times w_s \times 3}$ sampled from the input video $\mathcal{V}_{:T}$ at time $T$ 
and a sequence of low-resolution frames $\mathcal{V}_{T - t:T} \in \mathbb{R}^{t \times h_f \times w_f \times 3}$ taken $t$ time-steps before time $T$. 
First, we extract high-resolution 2D features from the $I_T$ input with a DINOv2 model\cite{oquab2023dinov2}, obtaining a set of 2D image tokens $\Phi_{2D}(I_T)$ and a class token $C_I$ offering a global representation of the image. We also extract spatio-temporal 3D features from the $\mathcal{V}_{T - t: T}$ input with a TimeSformer model~\cite{bertasius2021space} in the form of video tokens $\Phi_{3D}(\mathcal{V}_T)$ and a class token $C_\mathcal{V}$ giving a global representation of the input clip. 
\vspace{1mm}
\noindent
\textbf{Frame-guided Temporal Pooling Attention (Figure~\ref{fig:encoder}(a)):}
While the overall video tokens provide a spatio-temporal representation of the input video, STA predictions need to be aligned to the spatial location of the last video frame. The frame-guided temporal pooling attention maps video tokens to the spatial reference system of the last video frame, compressing the 3D representation obtained by the TimeSformer to a 2D one. 
The 3D video tokens $\Phi_{3D}(\mathcal{V}_{T-t:T})$ are mapped to 2D pooled video tokens denoted as $\Phi_{3D}^{2D}(\mathcal{V}_{T-t:T})$ adopting a residual cross-attention mechanism. 
Specifically, we compute query vectors from last-frame video tokens $\Phi_{3D}(\mathcal{V}_T)$ with a linear projection $W_{Q}$, while key and value vectors are computed from the overall video tokens $\Phi_{3D}(\mathcal{V}_{T-t:T})$ using the $W_K$ and $W_V$ linear projection layers. 
We obtain pooled video tokens with a residual multi-head attention ($A$) layer as follows:
\begin{equation}
\small
\Phi_{3D}^{2D}(\mathcal{V}_{T-t:T}) = \Phi_{3D}(\mathcal{V}_T) + A\big(
\underbrace{\Phi_{3D}(\mathcal{V}_{T}) W_{Q}}_{queries}, 
\underbrace{\Phi_{3D}(\mathcal{V}_{T - t:T})W_{K}}_{keys}, 
\underbrace{\Phi_{3D}(\mathcal{V}_{T - t:T})W_{V}}_{values}\big)    
\end{equation}

\noindent
Used as queries, last-frame tokens guide an adaptive temporal pooling that summarizes the spatio-temporal feature map computed by the TimeSformer model and maps it to the 2D reference space of the last observed frame. The residual connection facilitates learning and lets the attention mechanism focus on enriching last-frame tokens with video tokens.

\vspace{1mm}
\noindent
\textbf{Dual Image-Video Attention fusion (Figure~\ref{fig:encoder}(b)):}
Image tokens $\Phi_{2D}(I_T)$ and pooled video tokens $\Phi_{3D}^{2D}(\mathcal{V}_{T-t:T})$ are spatially aligned, but carry different information, with image tokens encoding fine-grained visual features and video tokens encoding scene dynamics.
This module adopts a residual dual cross-attention that aims to enrich image tokens with scene dynamics information coming from video tokens through image-guided cross-attention and, vice versa, video tokens with fine-grained visual information coming from image tokens through video-guided cross-attention. 
Prior to forwarding image and video tokens to the multi-head cross-attention modules, these are summed with learnable positional embeddings to capture insightful spatial relationships and normalized through a Layer Norm. The residual image-guided cross-attention is as follows:

\begin{align}
\small
&[\tilde{\Phi}_{2D}(I_T), \tilde{C}_I] = [\Phi_{2D}(I_T),C_I] + \notag
\\
&A(\underbrace{[\Phi_{2D}(I_T),C_I]W_{Q}}_{queries}, \underbrace{[\Phi_{3D}^{2D}(\mathcal{V}_{T-t:T}),C_\mathcal{V}]W_{K}}_{keys},
\underbrace{[\Phi_{2D}^{3D}(\mathcal{V}_{T-t:T}),C_\mathcal{V}]W_{V}}_{values})
\end{align}



\noindent
where $[\cdot,\cdot]$ denotes concatenation along batch dimension, and $W_Q$, $W_K$, and $W_V$ are linear projection layers. After the multi-head attention layer, the refined image representation $[\tilde{\Phi}_{2D}(I_T), \tilde{C}_I]$ is passed through a residual MLP. The video-guided cross-attention works in a similar way to compute refined video tokens $\tilde{\Phi}_{3D}(\mathcal{V}_{T-t:T})$ and video class tokens $\tilde{C}_\mathcal{V}$, but queries are computed from video tokens while keys and values are computed from image tokens.

\vspace{1mm}
\noindent
\textbf{Feature Fusion and prediction head (Figure~\ref{fig:encoder}(c)-(e)):}
Refined image and video class tokens are summed to obtain the overall class token $C_T = \tilde{C}_{I} + \tilde{C}_{\mathcal{V}}$, a global representation of the input image-video pair (Figure~\ref{fig:encoder}(c)). Refined image tokens $\tilde{\Phi}_{2D}(I_T)$ are mapped to a multi-scale feature pyramid~\cite{lin2017feature} $P_{2D}(I_T)$ by rescaling $\tilde{\Phi}_{2D}(I_T)$ to multiple resolutions using bilinear interpolation\footnote{See the supplementary material for more details. \label{fn:repeated_feet}}, followed by a $3 \times 3$ convolution to compensate for interpolation artifacts.
Refined video tokens $\tilde{\Phi}_{3D}(\mathcal{V}_{T-t:T})$ are mapped to a feature pyramid $P_{3D}(\mathcal{V}_{T-t:T})$ in the same way.
The two feature pyramids are summed and passed through a 2D $3 \times 3$ convolution to obtain the fused feature pyramid $P_T$ (Figure~\ref{fig:encoder}(d)). We adopt the prediction head$^{\ref{fn:repeated_feet}}$ proposed in~\cite{ragusa2023stillfast} to obtain the final predictions $(\hat b_i, \hat n_i, \hat v_i, \hat \delta_i, \hat s_i)$. It is a modified version of the Faster-RCNN~\cite{girshick2015fast} that integrates specialized components for STA prediction. Note that while~\cite{ragusa2023stillfast} uses global average pooling to obtain a global representation of the scene, we naturally use the class token $C_T$ learned from the input image-video pair.

\section{Leveraging affordances for human behavior grounding}
While end-to-end STA architectures predict future human-object interactions from labeled data, in this section we show that it is beneficial to explicitly incorporate environment affordances based on linking functionally similar regions and interaction hotspots obtained from hand trajectories.


\subsection{Leveraging environment affordances}

Environment affordances \cite{nagarajan2020ego} refer to all potential interactions that can be performed in a given physical zone. Our intuition is that a robust representation of environment affordances, learned from the observation of human activities in egocentric video, encapsulates 
the interaction that the user is going to perform next. We first build an affordance database grouping the training videos according to their visual similarity in activity-centric zones. At inference time, we match a novel video $\mathcal{V}'$ to the most functionally similar zones in the affordance database, estimating the distribution of the affordable interactions in the new video. We use the nouns and verbs affordance distributions to refine the respective nouns and verbs probabilities predicted by STAformer.


\vspace{1mm}
\noindent
\textbf{Building the affordance database:}
We start extracting activity-centric zones from the training set following~\cite{nagarajan2020ego}.
We build positive and negative frame pairs labels by counting homography estimation inliers, evaluating temporal coherence, and computing visual similarity with a pre-trained ResNet-152.
A Siamese network $\mathbb{L}$ is then trained on these pairs and used to predict the probability $\mathbb{L}(I, I')$ that two frames $I$ and $I'$ belong to the same zone. We then process all frames in a video sequence with $\mathbb{L}$ to group video frames according to their visual similarity in different zones.$^{\ref{fn:repeated_feet}}$
Each zone $Z$ represents an activity-centric region composed of the group of visually similar images $I_i^Z$, their corresponding videos $\mathcal{V}_i^Z$, the associated narrations $\mathcal{T}_i^Z$ , sets of nouns $\mathcal{N}_i^Z$ and action verbs $\mathcal{A}_i^Z$ appearing at least once in the STA annotations of all images $I_i^Z$. This represents a sort of \textit{persistent memory} on how humans behave in each different environment.
We obtain a visual descriptor $Z^\mathcal{V}$ and a text descriptor $Z^\mathcal{T}$ for each zone $Z$ computing the average descriptors of videos within each zone: $Z^\mathcal{V} = \sum_{i = 1}^{|Z|} \mathrm{\Psi}_\mathcal{V}(\mathcal{V}_i^Z) / |Z|$,  $Z^\mathcal{T} = \sum_{i = 1}^{|Z|} \mathrm{\Psi}_\mathcal{T}(\mathcal{\mathcal{T}}_i^Z) / |Z|$. We adopt the dual encoder of EgoVLP-v2 \cite{pramanick2023egovlpv2} to extract video $\mathrm{\Psi}^\mathcal{V}(\mathcal{V}_i^Z)$ and text $\mathrm{\Psi}^\mathcal{T}(\mathcal{\mathcal{T}}_i^Z)$ descriptors.

\begin{figure}[t]
\centering
\includegraphics[width=0.99\textwidth]{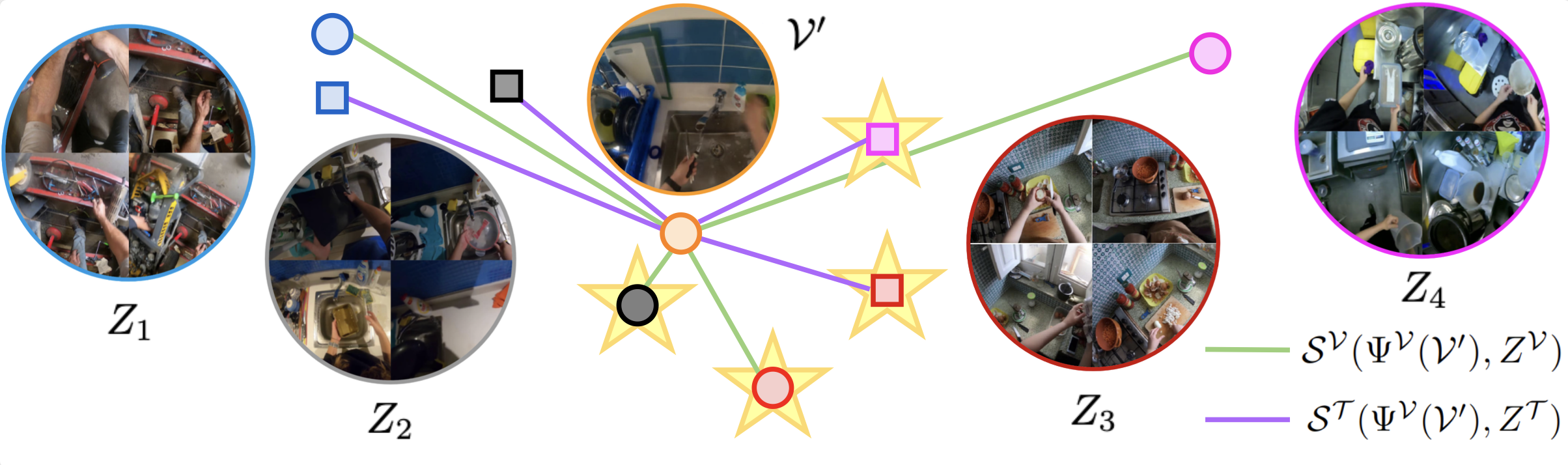} 
\\
\includegraphics[width=0.99\textwidth]{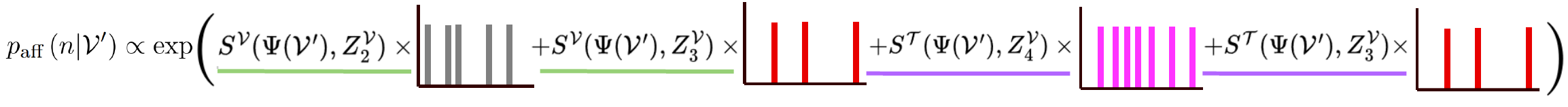}
\caption{\textbf{Cross-environment inference of affordances:} The input video $\mathcal{V}'$ is matched to the affordance database comparing its visual representation $\Psi^\mathcal{V}(\mathcal{V}')$ to the visual $Z^\mathcal{V}$ (\scalebox{2}{$\circ$}) and text $Z^\mathcal{T}$ ( $\square$) zone descriptors. The affordance noun probability $p_{\text{aff}}\left(n|\mathcal{V}'\right)$ is obtained by weighting the counts of nouns present in the top-2K nearest zones  (\scalebox{2}{$\star$}) according to the respective similarity $\mathcal{S}$. Example for K=2.}
\label{fig:aff_scheme}
\end{figure}

\vspace{1mm}
\noindent
\textbf{Inferring environment affordances:}
While~\cite{nagarajan2020ego} links similar activity-centric zones and trains a neural network to predict affordances directly from video, we found this approach suboptimal in our settings as we discuss in the results section. Instead, at inference time, we predict the nouns and verbs affordance distribution by matching a novel video $\mathcal{V}'$ to zones related to functionally similar environments in the affordance database. Since we can only extract a visual descriptor from the novel video, $\mathrm{\Psi}^\mathcal{V}(\mathcal{V}')$, we compute the visual cosine similarity $\mathcal{S}^\mathcal{V}(\mathrm{\Psi}^\mathcal{V}(\mathcal{V}'), Z^\mathcal{V})$ and the video-text cross cosine similarity $\mathcal{S}^\mathcal{T}(\mathrm{\Psi}^\mathcal{V}(\mathcal{V}'), Z^\mathcal{T})$ between the clip and each zone $Z$ in the database. Despite being visually dissimilar, the cross distance relates different locations with similar functionality that affords the same interaction (i.e, painting a wall in India or painting a canvas with watercolor in Spain both afford to dip the brush in the paint).

As illustrated in Figure~\ref{fig:aff_scheme}, we employ the K-Nearest Neighbour algorithm to identify the most similar zones to the given input $\mathcal{V}'$. We define the top-K visual zones $\mathcal{K}^{\mathcal{V}} =\{(Z^\mathcal{V}_1, S^\mathcal{V}_1), ..., (Z^\mathcal{V}_K,S^\mathcal{V}_K) \}$, where $S_k^\mathcal{V}$ is a shorthand notation for $S^\mathcal{V}_k(\Psi(\mathcal{V}'), Z_k^\mathcal{V})$, and the top-K narrative zones $\mathcal{K}^{\mathcal{T}}=\{(Z^{\mathcal{T}}_1, S^\mathcal{T}_1), ..., (Z^{\mathcal{T}}_K,S^\mathcal{T}_K) \}$. Combining both sets, $\mathcal{K} = \mathcal{K}^\mathcal{V} \cup \mathcal{K}^\mathcal{T} = \{ (Z_i, S_i) \}_{i=1}^{2K}$ yields a total of  $2K$ zones and their respective similarity scores, which we assume to share affordances with $\mathcal{V}'$. We then define the probability of each noun $p_{\text{aff}}\left(n|\mathcal{V}'\right)$ as an exponential distribution by weighting the noun appearance in each neighbouring zone $\mathcal{N}^{Z_i}$ according to the respective similarity $S_i$:
\begin{equation}
p_{\text{aff}}\left(n|\mathcal{V}'\right) \propto \exp (
\sum_{(Z_i, S_i) \in \mathcal{K}} S_i \cdot \mathbb{1}_{n \in \mathcal{N}^{Z_i}} )    
\end{equation}
We apply the same procedure to predict the verb distribution $p_{\text{aff}}\left(v|\mathcal{V}'\right)$.

\vspace{1mm}
\noindent
\textbf{STA predictions and environment affordances data fusion:}
Based on the environment affordances, we can predict probability distributions over \textit{possible} nouns $p_{\text{aff}}\left(n|\mathcal{V}\right)$ or verbs $p_{\text{aff}}\left(v|\mathcal{V}'\right)$ given \textit{past interactions in functionally similar zones}. Differently, the STA model will predict probability distributions of given nouns and verbs \textit{being the next interactions}  $p_{\text{sta}}\left(n|\mathcal{V}', I'\right)$ and $p_{\text{sta}}\left(v|\mathcal{V}', I'\right)$ directly from the input image-video pair, without explicitly considering the set of possible actions. 
We assume independence between the two predictions\footnote{In practice, we build the two models with different architectures and training objectives to make the dependence weak.} and perform data fusion by computing the unnormalized joint likelihoods:

\begin{equation}
\begin{split}
    p_{\text{fus}}(n|I', \mathcal{V}') &\propto p_{\text{aff}}\left(n|\mathcal{V}'\right) \cdot p_{\text{sta}}\left(n|\mathcal{V}', I'\right) \\
    p_{\text{fus}}(v|I', \mathcal{V}') &\propto p_{\text{aff}}\left(v|\mathcal{V}'\right) \cdot p_{\text{sta}}\left(v|\mathcal{V}', I'\right)
\end{split}
\end{equation}

\subsection{Leveraging interaction hotspots:}
\begin{figure}[t]
\centering
\includegraphics[width=\textwidth]{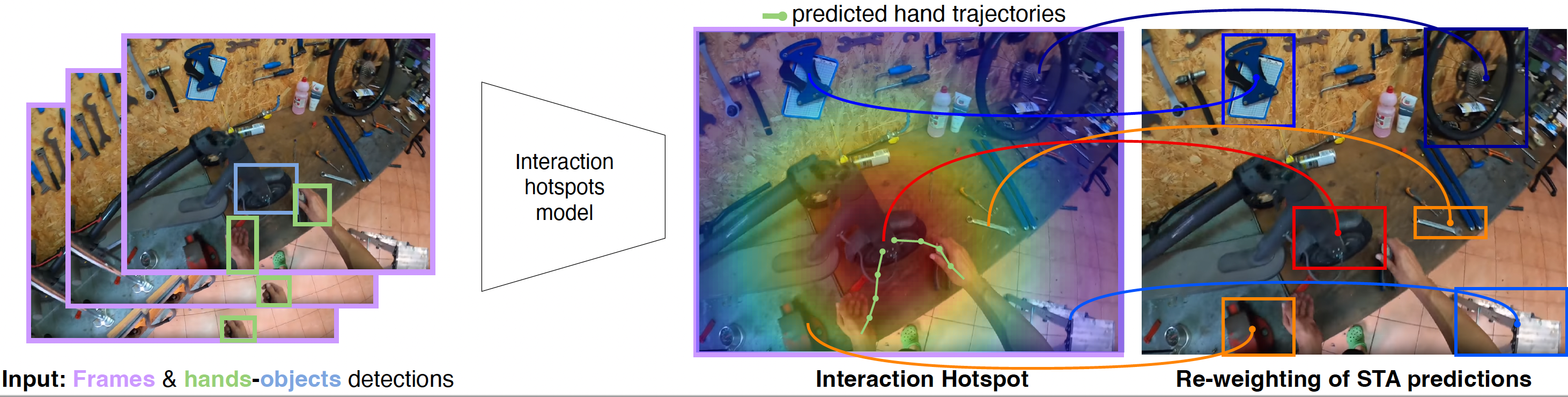}
\caption{\textbf{Refinement of confidence scores based on the interaction hotspots.} The interaction hotspot model observes frames, hands, and objects and forecasts a map encoding the probability of the interaction in each pixel. STA confidence scores are re-weighted based on the probability values at the bounding box coordinate centers, reducing confidence in false positive predictions falling far from the interaction hotspot.}
\label{fig:int_hots_teaset}
\end{figure}

While our affordance database gives us information on which objects (nouns) and interaction modes (verbs) are likely to appear in the current scene, it does not give us any information on \textit{where} the interaction will take place in the observed images.
As noted in previous works~\cite{liu2020forecasting,liu2022joint}, observing how hands move in egocentric videos can allow us to predict the interaction hotspot~\cite{liu2022joint,nagarajan2020ego}, a distribution over image regions indicating possible future interactions locations.
We exploit this concept and include a module to predict an interaction hotspot by observing frames, hands, and objects. As Figure~\ref{fig:int_hots_teaset} illustrates, we hence re-weigh the confidence scores $s_i$ of STA predictions according to the location of the respective bounding box centers in the predicted interaction hotspot, to reduce the influence of false positive detections falling in areas of unlikely interaction.

\vspace{1mm}
\noindent
\textbf{Inferring interaction hotspots} 
We base our interaction hotspot module on the work presented in ~\cite{liu2022joint} with some improvements.
First, we fine-tune the hand object detector presented in~\cite{shan2020understanding} on EGO4D-SCOD~\cite{grauman2022ego4d} annotations, rather than using it out-of-the-box.
Second, we extract stronger egocentric-aware frame features with the video part of the dual-encoder version of EgoVLP~\cite{pramanick2023egovlpv2} pre-trained on Ego4D~\cite{pramanick2023egovlpv2}, instead of using a ConvNet as in~\cite{liu2022joint}.\footnote{See supp. for more information on the interaction hotspot prediction module.} The model takes as inputs the features of the observed frames, besides the coordinates and features of both hands and pre-detected objects, and is trained to forecast the hand trajectory, from which it predicts a distribution over plausible future contact points.
Given the observed image-video pair $(I_T,\mathcal{V}_{T-t:T})$, the output of the model is a probability distribution over the spatial locations of $I_T$ indicating the probability of interaction of each pixel denoted as $p_{ih}(x, y |  I_T, \mathcal{V}_{T - t:T})$. 


\vspace{1mm}
\noindent
\textbf{Fusing STA predictions with interaction hotspots:}
We exploit the interaction hotspots to refine the predictions of the STA model, assuming that regions close to the predicted interaction hotspots are more likely to contain the next active objects. 
Given a predicted box $\hat b_i$, we re-weigh its related confidence score $\hat s_i$ according to the location of the bounding box center $(\hat c_i^x, \hat c_i^y)$ in the interaction hotspot as following: $\hat s_i \cdot p_{ih}(\hat c_i^x, \hat c_i^y|I_T, \mathcal{V}_{T - t:T})$. 

\section{Results}
We evaluate our approach on two challenging benchmarks: Ego4D~\cite{grauman2022ego4d} and EPIC-Kitchens~\cite{damen2018scaling}. Since no STA labels are available for EPIC-Kitchens, we extend and publicly release annotations to contribute a new benchmark for STA. 
We compare our model against different STA methods that either provide open-source implementations or report results in their papers\cite{grauman2022ego4d, sta2023quickstart, ragusa2023stillfast, pasca2023summarize, chen2022internvideo, thakur2023guided}.
We adopt standard Noun (N), Noun+Verb (N+V), Noun+time-to-contact (N+$\delta$) and Noun+Verb+time-to-contact (All) Top-5 mean Average Precision (mAP).

\subsection{Comparison with the state-of-the-art}
Tables~\ref{tab:mAP_v1}-\ref{tab:mAP_v2} report the results on the validation sets of Ego4D v1 and v2. Our method outperforms all previous approaches by wide margins, showing relative gains\footnote{{We compute the relative gain\% of $x$ concerning $y$ as $100 \cdot (\frac{x-y}{y})$.}} of up to $+45.0\%$ and $+42.1\%$ on v1 and v2 respectively when considering the mAP All measure. The significant improvements both in semantic, spatial, and temporal reasoning confirm the benefits of our two main contributions: STAformer and the integration of affordances.
The joined semantic generalization capacity of environment affordances and the spatial refinement of interaction hotspots make STAformer + AFF excel in the $N+V$ mAP, obtaining a gain of +58.9$\%$ and 47.6$\%$ on v1 and v2, respectively.
%
%

\begin{table}[t]
    \centering
    \begin{minipage}{0.49\linewidth}
    \centering
            \caption{Results in mAP on the validation split of Ego4D-STA v1. \textbf{Best results} in bold. Relative gain is with respect to \underline{second best}}
            \label{tab:mAP_v1}
        \scriptsize
        \begin{tabular}{|c|cccc|}
        \hline
        Model & N & N + V & N + $\delta$ & All \\ \hline
        FRCNN+SF~\cite{grauman2022ego4d} & 17.55 & 5.19 & 5.37 & 2.07 \\
        FRCNN+Feat.~\cite{sta2023quickstart} & \underline{22.01} & 5.52 & 5.54 & 1.78 \\
        StillFast~\cite{ragusa2023stillfast}  & 16.21 & 7.47 & 4.94 & 2.48 \\
        Transfusion~\cite{pasca2023summarize} & 20.19 & \underline{7.55} & \underline{6.17}& \underline{2.60} \\ \hline
        STAformer & 21.71 & 10.75 & 7.24 & 3.53 \\
        STAformer + AFF& \textbf{24.36} & \textbf{12.00} & \textbf{7.66}  & \textbf{3.77} \\
        \hline
        Gain (rel $\%$) & \textcolor{ForestGreen}{+10.6} & \textcolor{ForestGreen}{+58.9} & \textcolor{ForestGreen}{+24.2} & \textcolor{ForestGreen}{+45.0} \\\hline
        \end{tabular}
    \end{minipage}
    \hfill
    \begin{minipage}{0.47\linewidth}
    \centering
            \caption{Results in mAP on the validation split of Ego4D-STA v2.  \textbf{Best results} in bold. Relative gain is with respect to \underline{second best}.}
            \label{tab:mAP_v2}
        \scriptsize
        \begin{tabular}{|c|cccc|}
            \hline
            Model & N & N + V & N +$\delta$ & All \\ \hline
            FRCNN+SF~\cite{grauman2022ego4d} & 21.00 & 7,45 & 7.07 & 2.98 \\
            InternVideo~\cite{chen2022internvideo} & 19.45 & 8.00 & 6.97 & 3.25 \\
            StillFast~\cite{ragusa2023stillfast} & 20.26 & 10.37 & 7.26 & 3.96 \\
            GANO v2~\cite{thakur2023guided} & \underline{20.52} & \underline{10.42} & \underline{7.28} & \underline{3.99} \\ \hline
            STAformer  & 24.85 & 13.45 & 7.41 & 4.90 \\
            STAformer+AFF & 27.03 & 14.36 & 8.72 & 5.04 \\
            STAformer+MH & 27.51 & 14.68 & 9.63 & 5.50 \\
            STAformer+MH+AFF & \textbf{29.39} & \textbf{15.38} & \textbf{9.94} & \textbf{5.67} \\
            \hline
            Gain (rel $\%$) & \textcolor{ForestGreen}{+43.3}& \textcolor{ForestGreen}{+47.6}& \textcolor{ForestGreen}{+36.5}& \textcolor{ForestGreen}{+42.1}\\\hline
        \end{tabular}
       
    \end{minipage}
\end{table}

\begin{table}[t]
    \centering
    \begin{minipage}{0.49\linewidth}
    \centering
        \caption{Results in mAP on the test split of Ego4D-STA. T denotes training data.}
        \label{tab:mAP_test}
        \scriptsize
        \begin{tabular}{|c|c|cccc|}
        \hline
        Model  & T& N & N + V & N + $\delta$ & All \\ \hline
        FRCNN+SF.~\cite{grauman2022ego4d} & v1 & 20.45 & 6.78 & 6.17 & 2.45\\
        FRCNN+Feat.~\cite{sta2023quickstart} & v1 & 20.45 & 4.81 & 4.40 & 1.31 \\
         InternVideo ~\cite{chen2022internvideo}& v1& 24.60& 9.18& \underline{7.64}&3.40\\
 Transfusion~\cite{pasca2023summarize}& v1& \underline{24.69}& \underline{9.97}& 7.33&3.44\\
         StillFast \cite{ragusa2023stillfast}   & v1& 19.51& 9.95& 6.45& \underline{3.49}\\ \hline
         STAformer & v1 & 24.39 & 12.49& 7.54&4.03\\ 
         STAformer + AFF & v1& \textbf{26.52}& \textbf{13.15}& \textbf{7.78}&\textbf{4.06}\\ \hline 
         Gain (rel $\%$) & v1 & \textcolor{ForestGreen}{+7.4}& \textcolor{ForestGreen}{+32.1}& \textcolor{ForestGreen}{+1.8} & \textcolor{ForestGreen}{+13.1} \\\hline
         
         \hline
        StillFast \cite{ragusa2023stillfast}   &v2& 25.06 & 13.29 & 9.14 & 5.12 \\
        GANO v2 \cite{thakur2023guided}  &v2& 25.67 & \underline{13.60} & 9.02 & 5.16 \\
        Language NAO  &v2& \underline{30.43} & 13.45 & \underline{10.38} & \underline{5.18} \\ \hline
        STAformer &v2& 30.61 & 16.67 & 10.06 & 5.62 \\
        STAformer+AFF  & v2 & 32.39 & 17.38 & 10.26 & 5.70 \\\hline
        STAformer+MH &v2& 31.99 & 16.79 & 11.62 & 6.72 \\
        \begin{tabular}[c]{@{}c@{}}STAformer\\ +MH+AFF \end{tabular} & v2 & \textbf{33.50} & \textbf{17.25} & \textbf{11.77} & \textbf{6.75} \\\hline
        Gain (rel $\%$) & v2 & \textcolor{ForestGreen}{+10.1} & \textcolor{ForestGreen}{+28.3} & \textcolor{ForestGreen}{+13.4} & \textcolor{ForestGreen}{+30.3} \\\hline
        \end{tabular}
        
    \end{minipage}
    \hfill
    \begin{minipage}{0.47\linewidth}
    \centering
        \caption{Results in mAP on the validation split of EPIC-Kitchens. \textbf{Best results} in bold. Relative gain is with respect to \underline{second best}.}
        \label{tab:mAP_EK}

        \scriptsize
        \begin{tabular}{|c|cccc|}
            \hline
            Model & N & N + V & N + $\delta$ & All \\ \hline
            StillFast \cite{ragusa2023stillfast} & \underline{21.24} & \underline{12.41} & \underline{6.22} & \underline{3.28} \\ 
            \hline
            STAformer & 24.16 & 15.55 & 7.08 & 4.31\\
            STAformer + AFF & \textbf{26.19}& \textbf{16.49}& \textbf{7.18}& \textbf{4.69}\\
            \hline
            Gain (rel $\%$) & \textcolor{ForestGreen}{+23.3}& \textcolor{ForestGreen}{+32.8}& \textcolor{ForestGreen}{+15.4}& \textcolor{ForestGreen}{+42.9}\\\hline
        \end{tabular}  
        
    \end{minipage}
\end{table}
Table~\ref{tab:mAP_test} reports the results on the test split of Ego4D. Note that the test set is private, so we are only able to compare approaches showing test results in their papers. 
For fair comparisons, we report two settings with methods trained on v1 or v2 (a larger set also includes v1 annotations). Our method achieves consistent gains with respect to trained methods on v1, for instance, obtaining a $+13.1\%$ in mAP All and a $+32.1\%$ in N+V mAP. Smaller but consistent gains of $+7.4\%$ and $+1.8\%$ are obtained for N and N+$\delta$ mAP, respectively. We observe similar gains when training on v2, with a $+30.3\%$ mAP All, $+28.3\%$ N+V mAP, $+13.4\%$ mAP N+$\delta$ and $+10.1\%$ N mAP. 
It is worth noting that our approach also benefits from training on larger sets of data. Indeed, performance is improved in Table~\ref{tab:mAP_test} when training on v2, with respect to our model trained on v1, increasing from $4.06$ to $6.75$ mAP All and jumping from $26.52$ to $33.50$ N mAP due to the joining effect of the affordances and the multi-head attention. Table~\ref{tab:mAP_EK} finally reports the results on EPIC-Kitchens. Since this benchmark is new, we train the official implementation of StillFast~\cite{ragusa2023stillfast} on EPIC-Kitchens. Also in this case, our method achieves significant performance gains ranging from $+15.5\%$ in the case of N+$\delta$ mAP to $+42.9\%$ in mAP All.

\subsection{Ablation study}


\begin{table}[t]
\centering
\caption{Ablation study of the different components of STAformer on the v1-val split of Ego4D.  \faSnowflake Encoder frozen \faWrench  Encoder last-blocks finetuned \faCog  Full encoder trained.}
\label{tab:staformer_ablation}

\scriptsize
\resizebox{\textwidth}{!}{%
\begin{tabular}{|c|cccc|cccc|}
\hline
 Exp.& Image Encoder & Video Encoder & Temporal pooling & 2D-3D Fusion & N & N + V & N + $\delta$ & All \\ \hline
 \cite{ragusa2023stillfast}&R50 \faCog  & X3D \faCog  & Mean & Sum & 16.21 & 7.52 & 4.94 & 2.48 \\ \hline \hline
 A1&DINOv2  \faSnowflake & - & - & - & 17.48 & 8.64 & 5.20 & 2.52 \\
 A2&DINOv2  \faSnowflake & DINOv2  \faSnowflake & Mean & Sum & 15.82& 7.65 & 4.11 & 2.19 \\
 A3&DINOv2  \faSnowflake & X3D \faCog & Mean& Sum& 18.84 & 8.84 & 5.56 & 2.57 \\ \hline \hline
 B1&DINOv2 \faSnowflake & TimeSformer \faWrench & Mean & Sum & 16.67 & 8.38 & 5.16 & 2.63 \\
 B2&DINOv2  \faSnowflake & TimeSformer \faWrench & Conv & Sum & 17.36 & 8.75 & 6.05 & 2.94 \\ 
 B3&DINOv2  \faSnowflake & TimeSformer \faWrench & Frame-guided & Sum & 19.78 & 10.04 & 6.35 & 3.39 \\ \hline \hline
  C1&DINOv2  \faSnowflake& TimeSformer \faWrench & Frame-guided & Dual $I \leftrightarrow \mathcal{V}$ attn & 20.08& 10.21& 6.51& 3.47 \\
  C2&DINOv2  \faWrench & TimeSformer \faWrench & Frame-guided & Dual $I \leftrightarrow \mathcal{V}$ attn& 21.71& 10.75 & 7.24& 3.53 \\
  C3&DINOv2  \faWrench & TimeSformer \faWrench & Frame-guided & $I \xrightarrow{} \mathcal{V}$ c.attn & 20.01 & 10.04 & 5.80 & 3.01\\
  C4&DINOv2  \faWrench & TimeSformer \faWrench & Frame-guided & $\mathcal{V} \xrightarrow{} I$ c.attn& 20.12& 10.31& 6.30& 3.35\\
  C5&DINOv2  \faWrench & TimeSformer \faWrench & MH.Frame-guided & MH.Dual $I \leftrightarrow \mathcal{V}$ attn& \textbf{23.02} & \textbf{11.57}& \textbf{7.86}& \textbf{3.85}\\\hline
\end{tabular}%
}
\end{table}

\subsubsection{STAformer architecture:}
Table~\ref{tab:staformer_ablation} compares the performance effects of the main components involved in the STAformer architecture.
In experiment A1, we encode the image input with a pre-trained DINOv2 model~\cite{oquab2023dinov2} and discard the video, obtaining small gains with respect to the baseline~\cite{ragusa2023stillfast}.
While \cite{ragusa2023stillfast} fully trains both image-video encoders, the A1 version trains solely the STA prediction head and reflects the modelling capacity of DINOv2.
However, simply extracting per-frame DINOv2 features and performing mean temporal pooling (Exp. A2), decreases the performance and indicates the limits of DINOv2 in modeling video dynamics.
Using X3D \cite{feichtenhofer2020x3d}, a convolutional 3D CNN, as the video encoder in Exp. A3 leads to improvements with respect to A1 (e.g., 18.84 vs 17.48 N mAP and 5.56 vs 5.20 N+$\delta$), indicating the advantage of appropriately encoding video dynamics.


We compare different versions of temporal pooling in experiments B1-B3 of Table~\ref{tab:staformer_ablation} using a fine-tuned TimeSformer as video model.\footnote{We finetune the last three blocks of the model.} Computing the mean along the temporal dimension of the video features (exp. B1), leads to non-systematic gains compared to the image-only transformer baseline A1.
Using a convolutional module for temporal pooling (exp. B2) helps modeling temporal cues, improving N+$\delta$ mAP up to 6.05. However, the proposed frame-guided attention (exp. B3) achieves a joint spatio-temporal understanding of the video improving from 8.75 to 10.04 N+V mAP and from 2.94 to 3.39 All mAP.

Next, experiments C1-C5 of Table~\ref{tab:staformer_ablation} assess the contribution of the proposed Dual Image-Video Attention module for 2D-3D feature fusion.
Comparing experiments C1 vs. B3 shows small but consistent gains when dual image-video attention is used for fusion, as compared to simple sum fusion (20.08 vs. 19.78 N, 10.21 vs. 10.04 N + V, 6.51 vs. 6.35 N + $\delta$ and 3.47 vs. 3.39 All mAP), suggesting that it is beneficial to enrich image tokens with video tokens and vice versa for 2D-3D fusion. The effect is more significant when we finetune the last 3 blocks of the image encoder (exp C2), showing the benefits of adapting the generalistic feature space of DINOv2 to the egocentric perspective.
Using standard cross-attention layers only with image tokens ($I \to \mathcal{V}$ - C3) or video tokens ($\mathcal{V} \to I$ - C4) as queries, while still allowing to outperform simple sum fusion (B3), performs worse than the proposed dual image-video attention (C2), suggesting again the need to incorporate the refinement of both modalities.
Finally, incorporating multi-head attention on the temporal pooling and on the 2D-3D fusion (C5) produces a consistent improvement in all the metrics due to its ability to capture diverse patterns from multiple representation simultaneously.


\begin{table}[t] 
\caption{Ablation of the effect of environment affordances and interaction hotspots on StillFast and STAformer. Results on the Ego4D val v1 split in Top-5 mAP.}
\label{tab:priors}
    \begin{minipage}{0.49 \linewidth}
    \centering
\label{tab:priors_staformer}
    \scriptsize
    \begin{tabular}{|c|c|c|cccc|}
        
         \multicolumn{7}{c}{\textbf{Ours}} \\
         \hline
         STA model & E.AFF& I.H&  N &  N+V&  N+$\delta$ &  All\\ \hline
         StillFast&  \faTimes &  \faTimes&  16.20&  7.47&  4.94&  2.48\\
         StillFast&  \faCheck &  \faTimes&  \underline{18.44}&  \underline{8.46}&  \underline{5.47}&  \underline{2.85}\\
         StillFast&  \faTimes &  \faCheck &  17.82&  7.62&  5.05&  2.53\\
         StillFast&  \faCheck &  \faCheck &  \textbf{19.34}&  \textbf{8.58}&  \textbf{5.55}&  \textbf{2.95}\\ \hline
         \multicolumn{7}{c}{\textbf{EGO-TOPO~\cite{nagarajan2020ego}}} \\
        \hline
        STA model & E.AFF& I.H&  N &  N+V&  N+$\delta$ &  All\\ \hline
         StillFast&  \faCheck &  \faTimes&  14.92& 6.45& 4.01 & 2.14\\ \hline 
    \end{tabular}
    
    \end{minipage}
    \hfill
    \begin{minipage}{0.49\linewidth}
    \centering
    \label{tab:priors_stillfast}
    \scriptsize
    \begin{tabular}{|c|c|c|cccc|}
    
        \multicolumn{7}{c}{\textbf{Ours}} \\
         \hline
         STA model & E.AFF& I.H&  N &  N+V&  N+$\delta$ &  All\\ \hline
         STAformer & \faTimes & \faTimes & 21.71 & 10.75 & 7.24 & 3.53 \\ 
         STAformer & \faCheck & \faTimes & 23.55 & \underline{11.75} & \underline{7.55} & \underline{3.74} \\
         STAformer & \faTimes & \faCheck & \underline{23.63} & 11.38 & 7.51 & 3.66 \\
         STAformer & \faCheck & \faCheck & \textbf{24.36} & \textbf{12.00} & \textbf{7.66} &\textbf{ 3.77} \\ \hline
         \multicolumn{7}{c}{\textbf{EGO-TOPO~\cite{nagarajan2020ego}}} \\
    \hline
         STA model & E.AFF& I.H&  N &  N+V&  N+$\delta$ &  All\\ \hline
         
         STAformer & \faCheck & \faTimes & 17.21 & 8.45 & 5.32 & 2.64 \\  \hline
    \end{tabular}
    
    \end{minipage}
\end{table}

\begin{figure}[t]
\centering
\includegraphics[width=\textwidth]{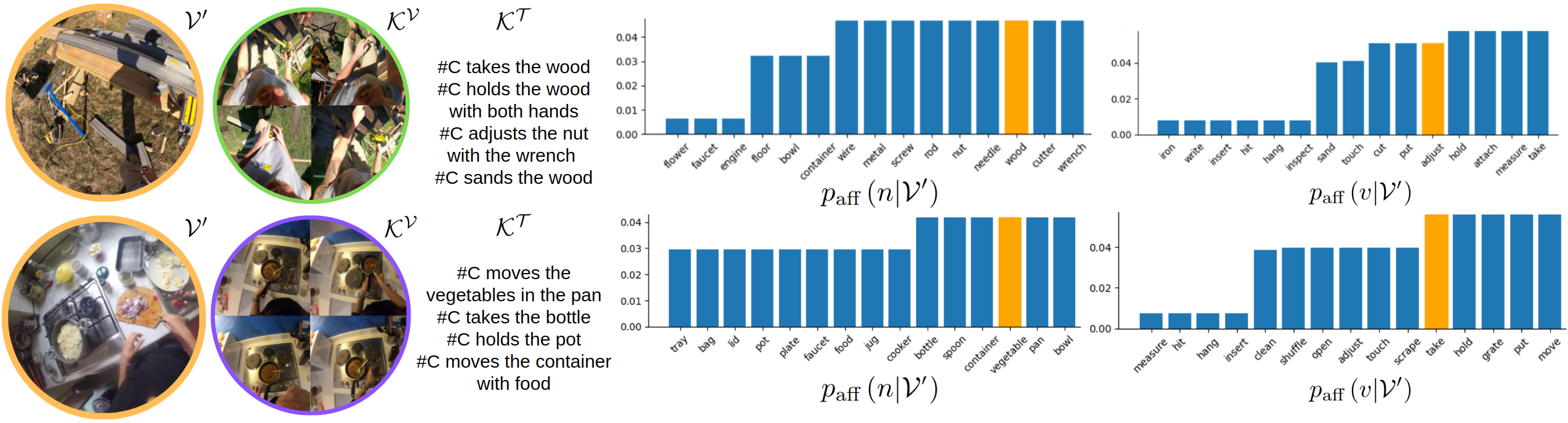} 
\caption{\textbf{Predicted environment affordances:} Linking across functionally similar environments ($\mathcal{K}^\mathcal{V}$, $\mathcal{K}^\mathcal{T}$) creates a robust affordance representation which captures the STA interaction. We show in orange the STA ground-truth label.}
\label{fig:aff_results}
\end{figure}

\begin{figure}[t]
    \centering
    \includegraphics[width=0.19\textwidth]{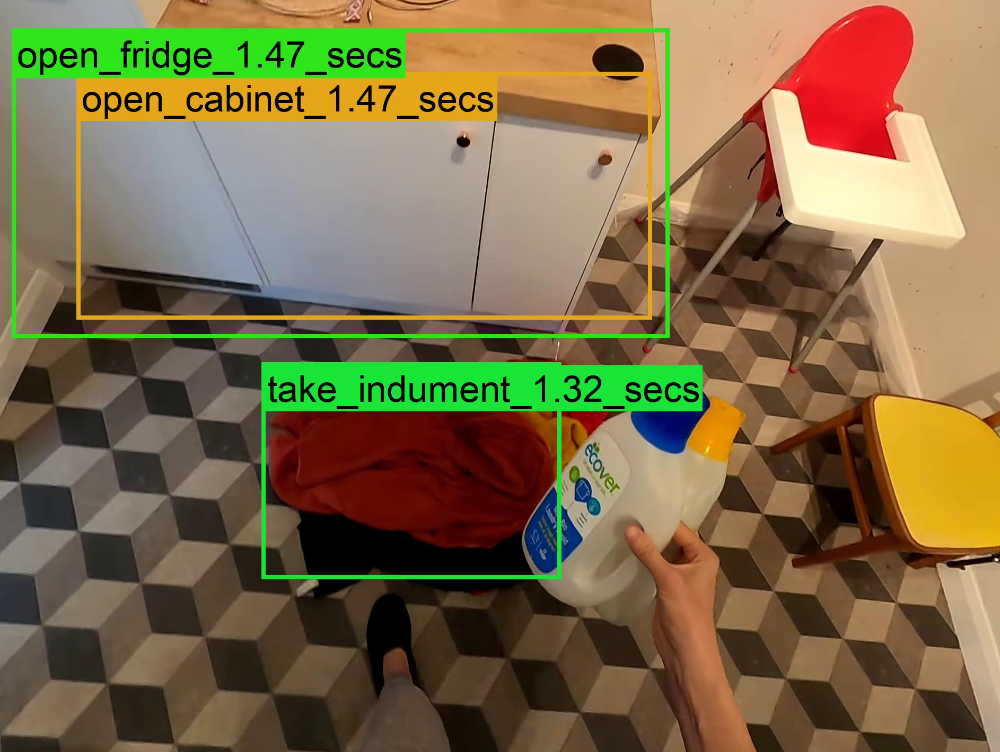}
    \hfill
    \includegraphics[width=0.19\textwidth]{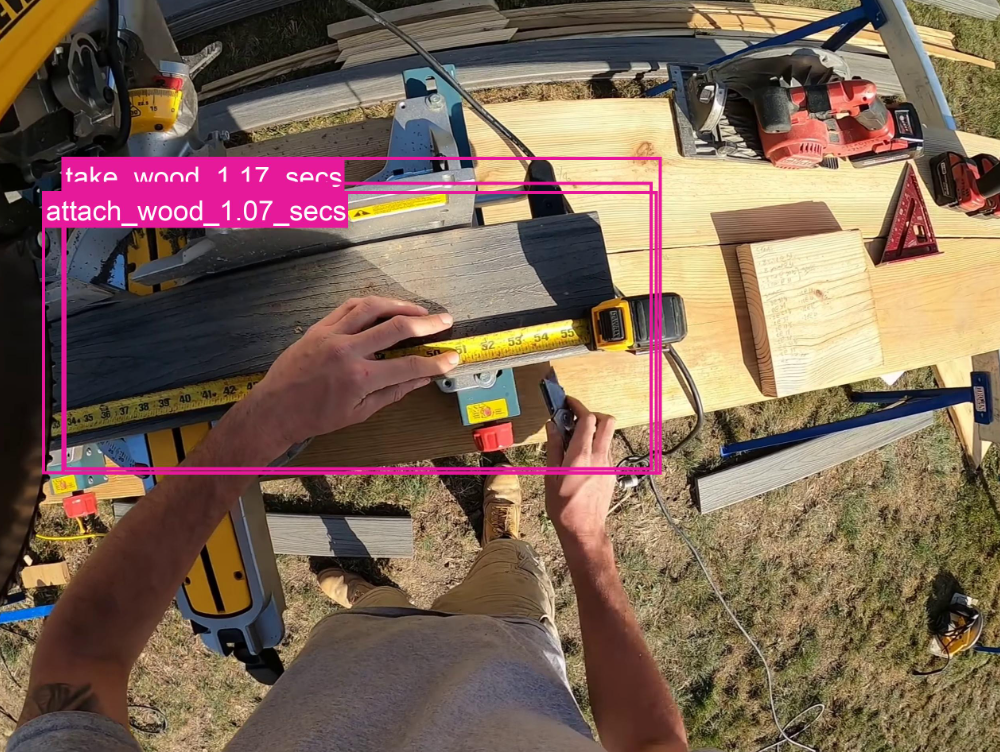}
    \hfill
    \includegraphics[width=0.19\textwidth]{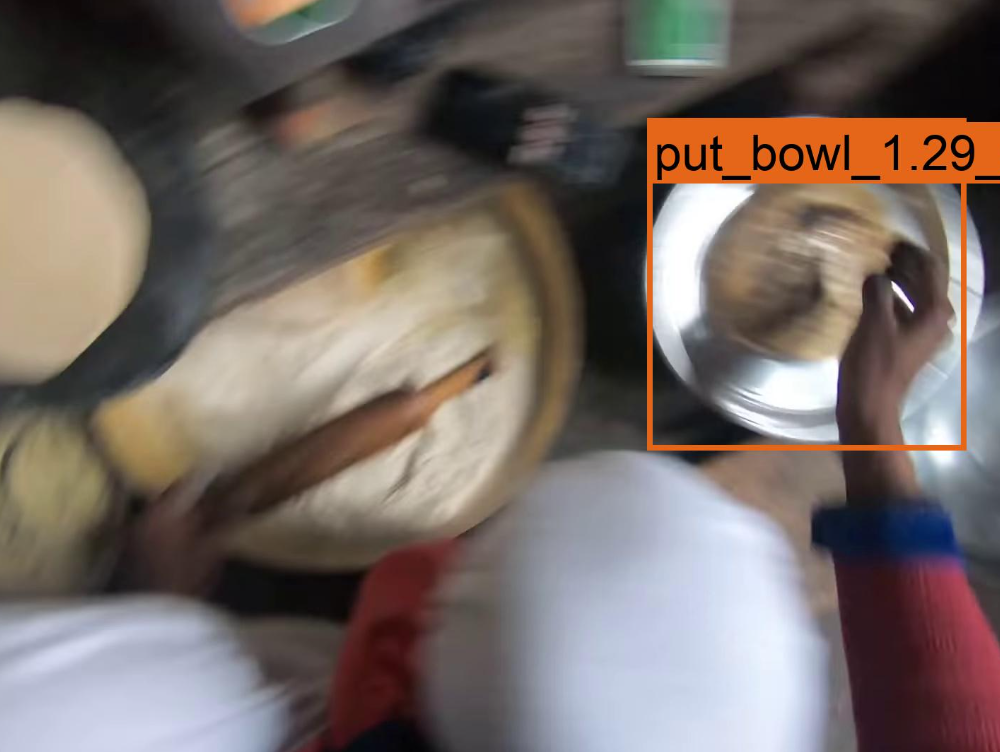}
    \hfill
    \includegraphics[width=0.19\textwidth]{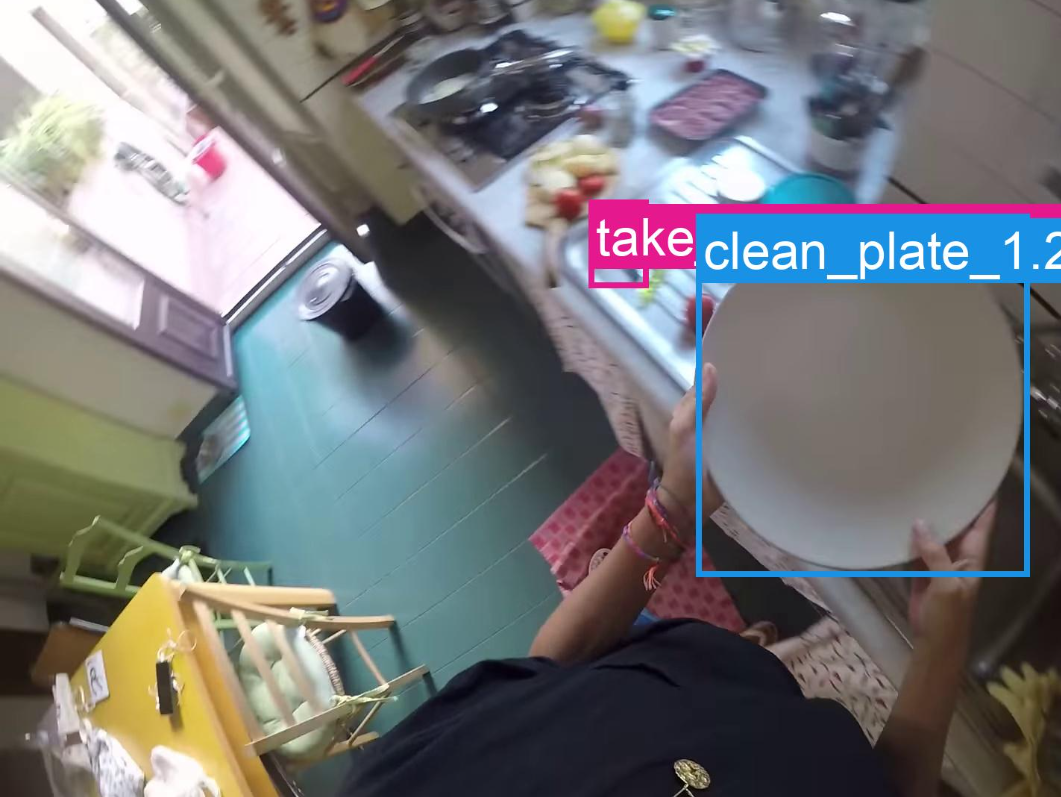}
    \hfill
    \includegraphics[width=0.19\textwidth]{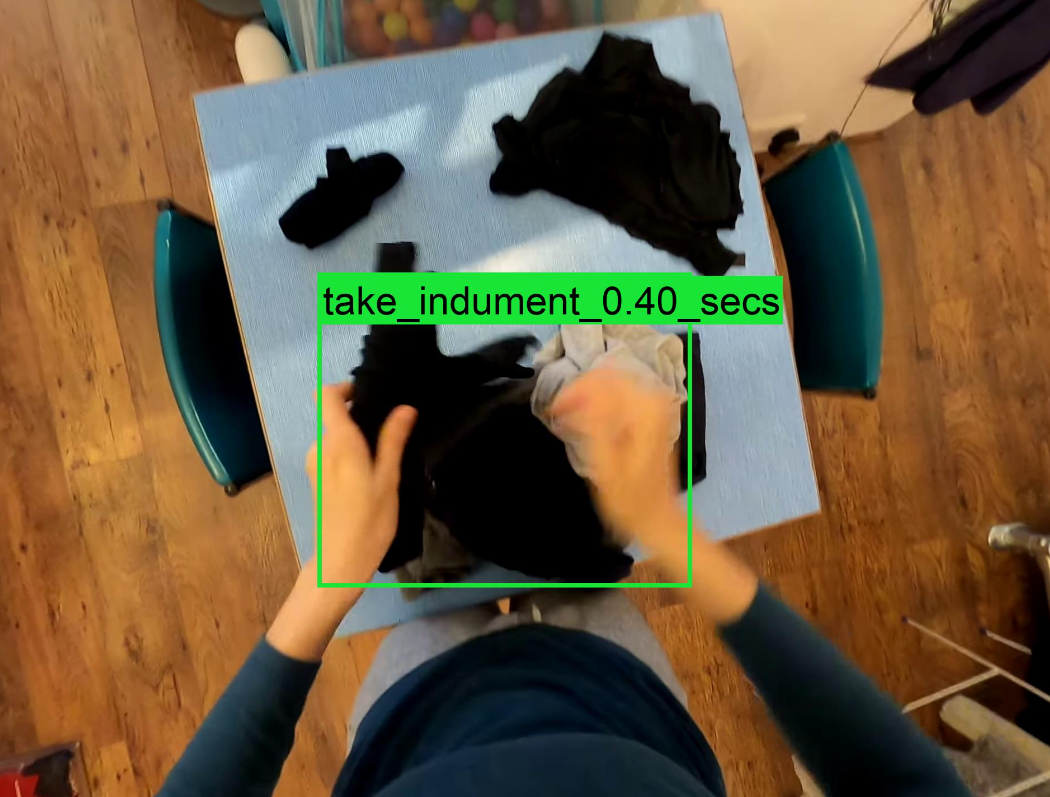}
    \\
    \includegraphics[width=0.19\textwidth]{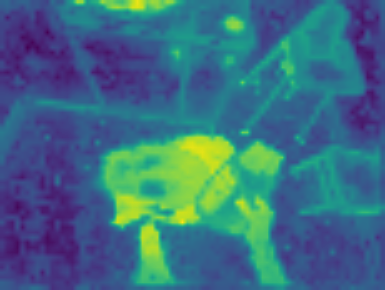}
    \hfill
    \includegraphics[width=0.19\textwidth]{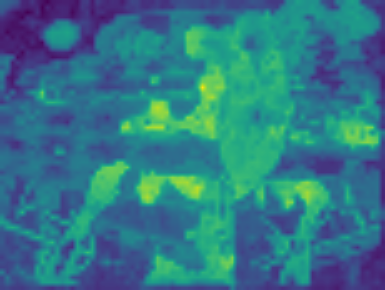}
    \hfill
    \includegraphics[width=0.19\textwidth]{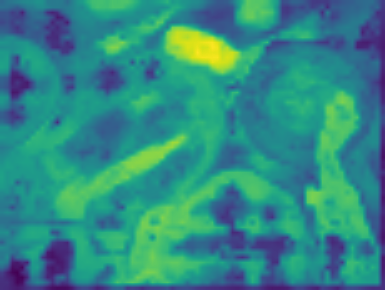}
    \hfill
    \includegraphics[width=0.19\textwidth]{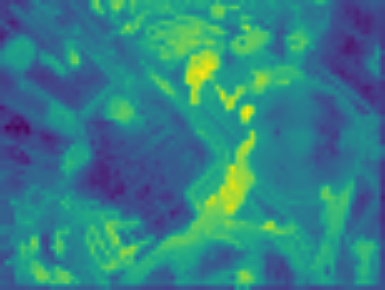}
    \hfill
    \includegraphics[width=0.19\textwidth]{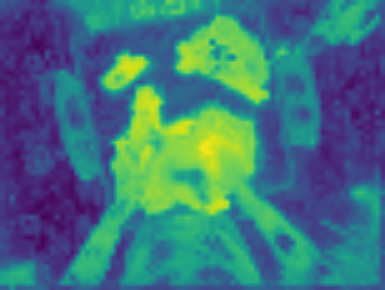}
    \\
    \centering
    \includegraphics[width=0.19\textwidth]{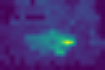}
    \hfill
    \includegraphics[width=0.19\textwidth]{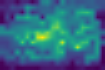}
    \hfill
    \includegraphics[width=0.19\textwidth]{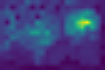}
    \hfill
    \includegraphics[width=0.19\textwidth]{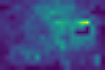}
    \hfill
    \includegraphics[width=0.19\textwidth]{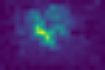}

    \caption{\textbf{Dual image-video attention maps, qualitative results.}
    Top to bottom: final predictions, attention map of pooled video tokens (queries) on image tokens (keys and values) and attention of image tokens (queries) on pooled video tokens (keys and values).
    Video tokens attend fine-grained object information from the high-resolution image; image features focus on objects which are important for future interactions.}
    \label{fig:imagenes_lado_a_lado}
\end{figure}


\vspace{1mm}
\noindent
\textbf{Leveraging affordances for STA:}
Table \ref{tab:priors} details the influence of environment affordances (E.AFF) and interaction hotspots (I.H), when integrated, separately and jointly, on the StillFast~\cite{ragusa2023stillfast} baseline and the proposed STAformer model, showing in both cases consistent improvements. 
The E.AFF module refines the nouns and verbs probabilities, obtaining significant gains in $N+V$ Top-5 mAP (8.46 vs. 7.47 in Stillfast and 11.75 vs. 10.75 in STAformer). However, training a NN classifier as in \cite{nagarajan2020ego} does not produce a useful distribution of the affordances for later fusion with the STA probabilties. Our intuition is that the NN overfits to the interactions in the scene which are more obvious, losing the generalist quality of our predictions across environments.
By re-weighing confidence scores based on the spatial prior provided by the interaction hotspots, the I.-H. module produces a general improvement in all the metrics (e.g., N mAP of 17.82 vs 16.20 in StillFast and 23.63 vs 21.71 in STAformer - mAP All of 2.53 vs 2.48 in StillFast and 3.66 vs 2.53 in STAformer). Combining environment affordances and hotspots brings significant improvements in both StillFast and STAformer. For instance, the proposed approach improves its N mAP from $21.72$ to $24.36$ and its All mAP from $3.53$ to $3.77$.

\subsection{Qualitative results}
Figure \ref{fig:aff_results} visualizes the nouns and verbs affordance distribution for two query videos, along with the closest zones with respect to visual appearance and narrations. Though the ground truth STA class is not the top predicted class, it is present in both the predicted verb and noun affordances, validating our hypothesis that scenes with close descriptive and visual similarity afford the same interaction.\footnote{We report more qualitative results in the supplementary material. \label{fn:results}}
Figure \ref{fig:imagenes_lado_a_lado} reports attention maps produced within the dual image-video attention module and final predictions (top). Video tokens attend fine-grained object information in the high-resolution image (middle), while image tokens attend scene dynamics in video features, which correspond to regions important for future interactions, such as moving hands or objects (bottom).$^{\ref{fn:results}}$

\section{Conclusions}
In this paper, we addressed the problem of Short-Term object-interaction Anticipation (STA). Our main contributions are STAformer, a novel attention-based architecture for STA, and the integration of affordances to ground STA predictions into human behavior.
Our results showcase the improvements given by the proposed architecture and affordance modules, which scores first on all splits of the challenging Ego4D and EPIC-Kitchens benchmarks.
We also detailed the contribution of each individual component through ablations and showed that the integration of affordances is beneficial also to other STA architecture besides the proposed one.
We will release the code and all the material, hoping that our work will enable future research in the area.


\section{Acknowledges}

Research at University of Catania has been supported by the project Future Artificial Intelligence Research (FAIR) – PNRR MUR Cod. PE0000013 - CUP: E63C22001940006. Research at the University of Zaragoza was supported by the Spanish Government (PID2021-125209OB-I00, TED2021-129410B-I00) and the Aragon Government (DGA-T45\textunderscore23R).

%
%



\bibliographystyle{splncs04}
\bibliography{main}

\end{document}